\documentclass[letterpaper]{article}
\pdfoutput=1
\usepackage{aaai25}
\usepackage{times}
\usepackage{helvet}
\usepackage{courier}
\usepackage[hyphens]{url}
\usepackage{graphicx}
\usepackage{natbib}
\usepackage{caption}
\usepackage{algorithm}
\usepackage{algorithmic}
\usepackage{newfloat}
\usepackage{listings}
\usepackage{booktabs}
\usepackage{multirow}
\usepackage{amsmath}
\usepackage{subcaption}

\DeclareCaptionStyle{ruled}{labelfont=normalfont,labelsep=colon,strut=off}
\floatstyle{ruled}
\newfloat{listing}{tb}{lst}{}
\floatname{listing}{Listing}

\title{MotifGPL: Motif-Enhanced Graph Prototype Learning for Deciphering Urban Social Segregation}
\author {
    Tengfei He\textsuperscript{\rm 1},
    Xiao Zhou\textsuperscript{\rm 1}\thanks{corresponding author}
}
\affiliations {
    \textsuperscript{\rm 1}Gaoling School of Artificial Intelligence, Renmin University of China\\
    hetengfei121@ruc.edu.cn, xiaozhou@ruc.edu.cn
}

\begin{document}

\maketitle

\begin{abstract}
Social segregation in cities, spanning racial, residential, and income dimensions, is becoming increasingly diverse and severe. As urban spaces and social relations grow more complex, residents in metropolitan areas experience varying levels of social segregation. If not promptly addressed, this could lead to an increase in crime rates, heightened social tensions, and other serious social issues. Effectively quantifying and analyzing the structures within urban spaces and resident interactions has become crucial for addressing these segregation issues. Previous studies have mainly focused on surface-level indicators of urban segregation, lacking comprehensive analyses from the perspectives of urban structure and mobility. This limitation fails to capture the full complexity of current segregation phenomena. To address this gap, we propose a framework named \textbf{Motif}-Enhanced \textbf{G}raph \textbf{P}rototype \textbf{L}earning (\textbf{MotifGPL}). The MotifGPL framework comprises three key modules: prototype-based graph structure extraction, motif distribution discovery, and urban graph structure reconstruction. Specifically, we use graph structure prototype learning to extract significant prototypes from both the urban spatial graph and the origin-destination graph, incorporating key urban attributes such as points of interest, street view images, and flow indices. To enhance interpretability, the motif distribution discovery module innovatively matches each prototype with similar motifs, which represent simpler graph structures reflecting local patterns. Finally, we use the motif distribution results to guide the reconstruction of the two graphs. This model facilitates a detailed exploration of urban spatial structures and resident mobility patterns, allowing us to identify and analyze the key motif patterns that influence urban social segregation, guiding the reconstruction of urban graph structures for lower segregation. Extensive experimental results demonstrate that MotifGPL can effectively reveal the key motif patterns influencing urban social segregation and provide robust guidance for mitigating this phenomenon.
\end{abstract}

\begin{links}
\link{Code}{https://github.com/tengfeihe/MotifGPL}
\end{links}

\section{Introduction}

Segregation refers to the differentiation of individuals from different population groups in either spatial or social dimensions, a phenomenon influenced by complex social contexts of environments \cite{oka2019segregation}. Residential segregation is a common phenomenon in global urban development. Previous studies have suggested a strong connection between residential segregation and issues such as disparities in educational distribution \cite{quillian2014does, owens2023little}, unequal employment opportunities \cite{bursell2023making}, and the uneven allocation of public infrastructure, including healthcare services \cite{white2012elucidating, seewaldt2023residential}. These imbalances caused by residential segregation can exacerbate tensions between residents, potentially leading to collective conflicts and regional unrest.

As urbanization advances, the growing complexity and diversity of urban areas are expanding social segregation beyond residential areas, leading to new forms of separation in social and economic realms \cite{lens2016strict}. 
Currently, income segregation is emerging as a significant issue in metropolises, restricting interactions among different groups. As \citeauthor{florida2017new} \shortcite{florida2017new} points out, rising income segregation strengthens social boundaries, reducing interactions between social classes and deepening societal divides.
These different forms of segregation reinforce each other, creating multi-layered segregation within cities.
Without effective policy interventions, these patterns of segregation may reinforce themselves, preventing sustainable development. Therefore, a better understanding of social segregation and the implementation of effective solutions are crucial for promoting urban development.

To explore and understand urban social segregation, sociologists have traditionally relied on static socioeconomic data to develop comprehensive indicators reflecting segregation levels. \citeauthor{massey1988dimensions} \shortcite{massey1988dimensions} initially categorized urban segregation into five primary dimensions, which \citeauthor{brown2006spatial} \shortcite{brown2006spatial} later distilled into two main dimensions: concentration-evenness and clustering-exposure. Building on this, \citeauthor{kwan2013beyond} \shortcite{kwan2013beyond} introduced an additional accessibility indicator, utilizing these three metrics to assess segregation levels. The advent of digitalization in urban environments has enriched the datasets available for segregation studies, offering a more dynamic perspective. Researchers have begun employing mobility data to track individual movement patterns within cities and to examine spatial accessibility \cite{chen2018understanding}. Furthermore, \citeauthor{moro2021mobility} \shortcite{moro2021mobility} demonstrated that severe income segregation can persist even in geographically proximate areas, highlighting the necessity of accounting for both urban spatial configurations and mobility patterns in segregation assessments.

Alongside the expansion of data types, innovative methodologies have been incorporated into segregation research. \citeauthor{sousa2022quantifying} \shortcite{sousa2022quantifying} utilized urban network structures to model racial segregation and employed graph random walks to explore the spatial diversity of racial distributions. Similarly, \citeauthor{zhang2021discovering} \shortcite{zhang2021discovering} leveraged mobility data and community detection algorithms to investigate income-economic segregation, analyzing movement patterns across different socioeconomic strata. These advancements not only broaden the scope of data used but also introduce new analytical techniques for more effectively dissecting complex social phenomena.

While extensive research has focused on quantifying social segregation using established metrics, these studies often lack depth in their interpretative analyses. Existing research frequently limits the use of mobility and urban spatial data to descriptive statistics, without exploring their potential to reveal deeper insights into segregation dynamics. Emerging technologies like deep learning are rarely applied to the study of social segregation, primarily due to a traditional reliance on established statistical methods. This underutilization of advanced technologies restricts the depth of analysis, preventing a more thorough exploration of the complex causes and patterns that characterize urban social segregation. Embracing these technologies could enhance the interpretability of segregation studies, offering deeper insights into the underlying mechanisms and facilitating a more comprehensive understanding of both the immediate and systemic factors driving segregation. This approach not only enriches the analytical landscape but also strengthens the potential for developing more effective interventions.

To address existing challenges, we introduce a framework named \textbf{Motif}-Enhanced \textbf{G}raph \textbf{P}rototype \textbf{L}earning (\textbf{MotifGPL}), incorporating interpretable deep learning methods to analyze social segregation through urban graph structures and mobility patterns. MotifGPL consists of three main components: prototype-based graph structure extraction, motif distribution discovery, and urban graph structure reconstruction. Our research leverages multimodal social data, which include socioeconomic indicators, population flow indices, \textbf{points of interest} (\textbf{POIs}), street view images, urban spatial graph, and \textbf{origin-destination} (\textbf{OD}) graph among other diverse information sources. Specifically, we first map this data onto nodes and edges of the urban graph to accurately reflect urban complexity. We then employ a prototype network graph feature extractor to encode the urban graph and learn prototype vectors that represent various socioeconomic states, capturing the essential features of social segregation. Through motif distribution discovery techniques, we delve into the fundamental structural patterns of the urban graph, essential for deciphering the micro-mechanisms of social segregation. 
We focus on the segregation index \cite{moro2021mobility} as a supervisory signal for our model's learning, enhancing interpretability and providing new insights into the dynamics of segregation. This metric aids in identifying and analyzing local structures within the urban graph and helps quantify the extent and patterns of social segregation. 
Our experimental results demonstrate that MotifGPL effectively identifies the key factors influencing urban social segregation at the motif pattern level, offering robust support for strategies aimed at reducing this issue. Our contributions are summarized as follows:

\begin{itemize}
    \item We propose a framework that analyzes social segregation by focusing on urban structures and mobility, effectively addressing the complexities of urban social segregation.
    \item Our model provides significant interpretability through motif distribution discovery, offering a deeper understanding of how urban structures contribute to social segregation and demonstrating a clear link between theoretical models and real-world urban dynamics.
    \item Our model offers actionable strategies for urban planning by utilizing motif analysis, which can effectively reduce segregation and foster urban sustainable development.
\end{itemize}

\section{Related Work}

\subsubsection{Social Segregation Assessment}
The concept of segregation, originally derived from sociological studies on racial segregation in cities, has had a profound impact on urban economies. For instance, segregation in the United States led to considerable economic disparities, including a \$3 billion loss for Black residents in Chicago, alongside skewed public resource allocation \cite{wang2018urban}. As urban dynamics have evolved, segregation now spans racial, residential, and income dimensions, intensifying as critical urban development challenges \cite{gottdiener2019new}.

Addressing urban segregation requires robust quantification of the issue. \citeauthor{massey1988dimensions} \shortcite{massey1988dimensions} initially categorized social segregation into five indicators: evenness, exposure, concentration, centralization and clustering. Later,  \citeauthor{brown2006spatial} \shortcite{brown2006spatial} simplified these to concentration and exposure for a more effective assessment. However, as urban functions and mobility patterns grow increasingly complex, these traditional indicators often fall short. \citeauthor{kwan2013beyond} \shortcite{kwan2013beyond} introduced accessibility as an additional dimension, providing a more comprehensive framework to assess segregation from both spatial and individual perspectives.

Existing methods typically assign a degree indicator to each block to reflect one aspect of segregation. While these methods provide a straightforward reflection of segregation distribution, they fail to uncover the structural information related to social segregation within urban spatial structures and population mobility.

\subsubsection{Graph Learning in Social Computing}
Modeling cities as graphs that incorporate edge information, rather than merely calculating statistical features of urban blocks, provides a richer representation of the city. Graph neural networks \cite{scarselli2008graph} are effective for embedding and predicting features in urban graph networks. Researchers \cite{huang2023learning, jin2023spatio, zou2024deep} use graph learning to analyze urban characteristics, dividing tasks into urban graph embedding \cite{li2024urban, yan2024urbanclip} and graph representation learning \cite{li2023urban, jin2023spatio, khoshraftar2024survey}. These tasks facilitate aligning diverse data—from POIs \cite{huang2023learning} to street view images \cite{yong2024musecl} and mobility data \cite{huang2023reconstructing}—into a unified feature space, which is then used to derive node attributes for downstream applications such as crime rate prediction \cite{xu2024cgap}, real estate price forecasting \cite{brimos2023explainable} and light pollution prediction\cite{zhang2024causally}.

Focusing on social segregation, \citeauthor{he2020human} \shortcite{he2020human} utilized social graph data to analyze individual mobility patterns and observe residential segregation from a personal perspective. \citeauthor{yabe2023behavioral} \shortcite{yabe2023behavioral} examined the effects of the COVID-19 pandemic on income segregation in U.S. cities using mobile signaling and statistical data. Additionally, graph community detection algorithms \cite{zhang2021discovering, cavallari2017learning} have been applied to analyze urban income segregation, offering new insights for segregation studies.

\subsubsection{Interpretability in Social
Computing}
In the field of graph deep learning, there is a growing emphasis on the importance of model interpretability, particularly for addressing complex sociological issues such as social segregation. Understanding the reasoning behind model outputs is crucial for effectively tackling these societal challenges. \citeauthor{kakkad2023survey} \shortcite{kakkad2023survey} categorizes interpretable graph models into post-hoc \cite{baldassarre2019explainability, ying2019gnnexplainer, zhang2022gstarx} and self-explainable \cite{zhang2022protgnn, seo2024interpretable, chen2024tempme} models. Post-hoc explanations analyze a trained model's weights to explain predictions, while self-explainable models integrate information or structural constraints during training to provide inherent explanations.
Incorporating deep learning into sociological studies enhances the uncovering of hidden information, thus improving our understanding of intricate social issues. \citeauthor{fan2023interpretable} \shortcite{fan2023interpretable} introduced an interpretable framework that elucidates complex interactions among urban variables to address income inequality. \citeauthor{tang2023explainable} \shortcite{tang2023explainable} added interpretability to spatiotemporal GNNs, enhancing predictions of traffic flow and urban space. \citeauthor{zhou2024explainable} \shortcite{zhou2024explainable} developed an interpretable model for simulating crowd movements, offering insights into urban dynamics. Similarly, \citeauthor{ding2024understanding} \shortcite{ding2024understanding} employed motif discovery algorithms to provide a deeper understanding of the factors influencing tourist attractions. 

Although these studies provide valuable tools for exploring urban issues, there is currently a lack of research that combines motif discovery with self-explainable graph learning frameworks to tackle urban problems.

\section{Preliminaries}

In this section, we introduce the urban graph structure, regional attribute data types, and the calculation of social segregation indices, outlining how these elements are integrated to effectively analyze urban social segregation.

\subsubsection{Urban Graph}
Cities consist of distinct blocks, each with unique geographical locations, traffic patterns, and commercial structures, which together form the city's fundamental geographic structure. Within this space, the movement of residents and the exchange of information between blocks create an urban OD graph. Thus, a city is represented by a graph $\mathcal{G}=(\mathcal{G}_s, \mathcal{G}_o)$, where $\mathcal{G}_s=(\mathcal{V}, \mathcal{E}_s)$ is the geographic spatial graph and $\mathcal{G}_o=(\mathcal{V}, \mathcal{E}_o)$ denotes the residents' OD graph. Here, $\mathcal{V}$ represents the city blocks, $\mathcal{E}_s$ the set of geographic proximity edges, and $\mathcal{E}_o$ the flow edges depicting resident movements. The adjacency matrices $A_s=(a^s_{ij} \in \{0,1\}), \forall i,j \in [1, \parallel\mathcal{V}\parallel ]$ and $A_o=(a^o_{ij} \in \{0,1\}), \forall i,j \in [1, \parallel\mathcal{V}\parallel ]$ correspond to $\mathcal{G}_s$ and $\mathcal{G}_o$, respectively.

\subsubsection{Region Attributes}
Regional attributes encompass both the geographical and social characteristics of city blocks. Street view data capture the geographical features of a block, the regional pedestrian flow index highlights its importance within the urban mobility framework, and POIs data reflect a block's role in the urban functional layout. Socioeconomic indicators represent the social characteristics of residents and measure the degree of social segregation.
For each block $v_i \in \mathcal{V}$, we define $X^{SV}_i$ as the street view data, $X^{FL}_i$ as the pedestrian flow index, $X^{POI}_i$ as the count of various POIs, and $X^{SE}_i$ as the socioeconomic indicators. The metrics $X^{SV}, X^{FL}, X^{POI},$ and $X^{SE}$ collectively represent these attributes for the entire block set $\mathcal{V}$.

\subsubsection{Network Motifs}
Network motifs are recurrent local connection patterns in complex network structures, serving as fundamental components that encapsulate the network's architecture. In prototype learning, each prototype $\mathbf{p}_i$ in the matrix $\mathbf{P}$ is associated with a motif pattern distribution $\mathbf{m}_i$, derived from the local graph structure most similar to $\mathbf{p}_i$. This distribution $\mathbf{m}_i$ reflects the local structural information of the node subset $\mathcal{V}_i$ that closely resembles the prototype.

\subsubsection{Social Segregation Index}
We collected aggregated socioeconomic indicators for each block, including average income levels, average education levels, and the average age of residents. Following \citeauthor{moro2021mobility} \shortcite{moro2021mobility}, we used the segregation index to calculate social segregation within each block, defined as follows:
\begin{equation}
    S_i = \frac{c}{2(c-1)}\sum_k^c\left|\tau_{ci} - \frac{1}{c}\right|,
\end{equation}
where $\tau$ is the economic distribution (e.g., income, education, and age) and $c$ is the dimension.

This segregation index quantifies differences in segregation levels across blocks and categorizes them into two classes based on quantiles to identify blocks with social segregation, providing a foundation for analyzing how urban networks influence segregation.

\begin{figure*}[t]
\centering
\includegraphics[width=0.9\textwidth]{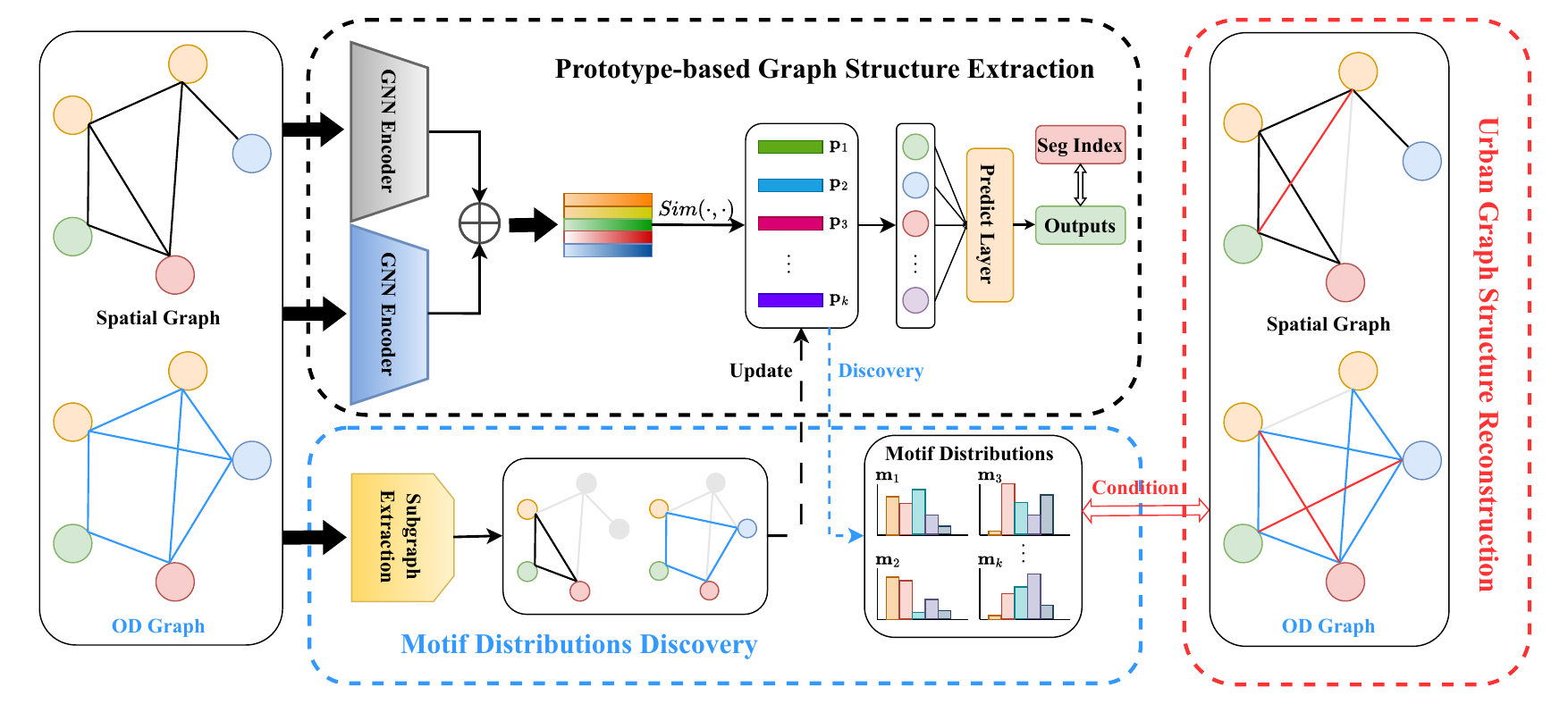} 
\caption{The Framework of Motif-Enhanced Graph Prototype Learning (MotifGPL).}
\label{fig1}
\end{figure*}

\section{Problem Statement}

Given an urban geographic spatial graph $\mathcal{G}_s=(\mathcal{V},\mathcal{E}_s)$ and an OD graph $\mathcal{G}_o=(\mathcal{V},\mathcal{E}_o)$ with their corresponding adjacency matrices $A_s$ and $A_o$, and node attributes $X=(X^{SV} \parallel X^{FL} \parallel X^{POI})$, we aim to use the degree of social segregation $\mathbf{d}^{SEG}$, derived from socioeconomic indicators $X^{SE}$, as a supervisory signal to learn prototypes $\mathbf{P}$ related to urban social segregation and their corresponding motif distribution $\mathbf{M}.$
The learning task is formalized as follows:
\begin{equation}
    F: \mathcal{D} \rightarrow M \in \mathrm{R}^{q \times d},
\end{equation}
where $\mathcal{D}=\langle \mathcal{G}_s, \mathcal{G}_o, A_s, A_o, X \rangle$, $q$ represents the number of prototypes, and $d$ denotes the dimension of the motif distribution. $F$ is a function that uses data from both graphs to compute motif distributions associated with the prototypes. Subsequently, we will utilize the motif distributions to guide the optimization of urban graph structures, aiming to reduce social segregation within cities.

\section{Methodology}

\subsection{Framework Overview}
Figure \ref{fig1} illustrates our framework for identifying local structures associated with social segregation from spatial and OD graphs. The model comprises three main components: the {\bf prototype-based graph structure extraction module}, the {\bf motif distribution discovery module}, and the {\bf urban graph structure reconstruction module}.

The prototype-based graph structure extraction module involves encoding multi-attribute and graph structure information, followed by prototype learning guided by social segregation degree indicators. The motif distribution discovery module then projects these prototypes onto the spatial and OD graphs to identify local node structures and extract motif distributions. Finally, the urban graph structure reconstruction module utilizes these motif distributions to optimize the urban spatial graph structure, with the goal of alleviating social segregation in cities.

\subsection{Prototype-based Graph Structure Extraction}
\subsubsection{Graph Encoder}
The urban spatial graph $\mathcal{G}_s$ considers the connectivity of urban areas based on geographical proximity. Following Tobler's first law of geography, we use the spatial graph to simplify inter-regional connections within cities, focusing on links between neighboring areas within a specified threshold. Conversely, the OD graph $\mathcal{G}_o$ represents connections based on population mobility.

Our model employs two feature extractors to independently encode $\mathcal{G}_s$ and $\mathcal{G}_o$ using adjacency matrices $A_s$ and $A_o$ and node attributes $X$. We utilize Graph Neural Networks (GNNs) as encoders to integrate graph structural information through a multi-layer message passing mechanism. The message passing formula is defined as follows:
\begin{equation}
    H^{(l+1)}_m = \sigma\left(\tilde{D}_m^{-\frac{1}{2}}\tilde{A}_m\tilde{D}_m^{-\frac{1}{2}}H_m^{(l)}W_m^{(l)}\right),
\end{equation}
where $m=\{s,o\}$ specifies encoders for $\mathcal{G}_s$ or $\mathcal{G}_o.$ $\tilde{A}_m = A_m + I_N$ includes self-loops, $\tilde{D}_m$ is the degree matrix, $\tilde{W}^{(l)}_m$ is the layer-specific weight matrix, and $\sigma$ typically refers to the ReLU activation function. $H^{(0)}=X$ represents the initial node attributes. To fully leverage the structural information from both graphs, we concatenate the outputs from each layer of the GNN encoders, resulting in a node latent space representation that blends features from both $\mathcal{G}_s$ and $\mathcal{G}_o$.

Following the encoding, the node attributes and graph structure are transformed into a latent space representation, denoted as $\mathbf{H}$. In the prototype learning layer, we maintain a fixed number ($N_{proto}$) of prototypes per class, capturing essential local structures. Similarity scores $s_{ij}$ between each node’s latent representation $\mathbf{h}_i$ and prototype $\mathbf{p}_j$ are calculated as follows:

\begin{equation}
    s_{ij} = \log \left(\frac{\parallel\mathbf{h}_i-\mathbf{p}_j\parallel^2_2 + 1}{\parallel\mathbf{h}_i-\mathbf{p}_j\parallel^2_2 + \epsilon}\right),
\end{equation}
where $\epsilon$ is a small constant for numerical stability. The similarity scores inform a fully connected layer with softmax activation to classify nodes.

\subsection{Motif Distribution Discovery}
\subsubsection{Prototype Projection}
Though prototype vectors are continuously optimized during training, they often lack intuitive interpretability. To address this, we developed a module that projects prototype vector $\mathbf{p}_i$ onto local subgraphs $\mathcal{G}_{sub}$ of nodes from category $k$ within the graph $\mathcal{G}$ during training. This projection enhances the interpretability of the prototype vectors by associating them with the local structures of specific node categories. For a given node $i$, with the representation $l_i$ denoting its local structure, the projection of prototype vectors is facilitated by the following process:
\begin{equation}
    \mathbf{p}_k = \mathop{\arg\min}\limits_{l_i \in L_i} \parallel l_i - \mathbf{p}_k \parallel_2,
\end{equation}
\begin{equation}
    L_i = \{f_{sub}(\mathcal{G}_{sub}^i); ~\mathcal{G}_{sub}^i \in \mathrm{Sub}(\mathcal{G}^i) ~\forall i ~\mathrm{s.t.}~ y_i = k\}.
\end{equation}
To efficiently classify nodes in large-scale graphs, we developed a random-walk-based local structure extractor. For each node $i$, this extractor generates $r$ random walk sequences, each of length $t$:
\begin{equation}
    s_{j+1} = \mathrm{Random}(\omega_j \mathcal{N}(s_{j})),
\end{equation}
\begin{equation}
    T^i \in \mathrm{R}^{r\times t}, \quad where\quad t^i_{k} = (s_0^k, s_1^k,\cdots, s_t^k),
\end{equation}
where $s_j$ denotes the currently selected node in the random walk, and $\omega_j$ represents the weight of the edge connecting to neighbors. $T^i$ represents the local graph structure of node $i$. Subsequently, we use a RNN to encode this local structure into latent space, matching the prototype's dimension, resulting in the encoded structure $l_i = \mathrm{RNNEncoder}(T^i)$.
\begin{figure}[t]
\centering
\includegraphics[width=\columnwidth]{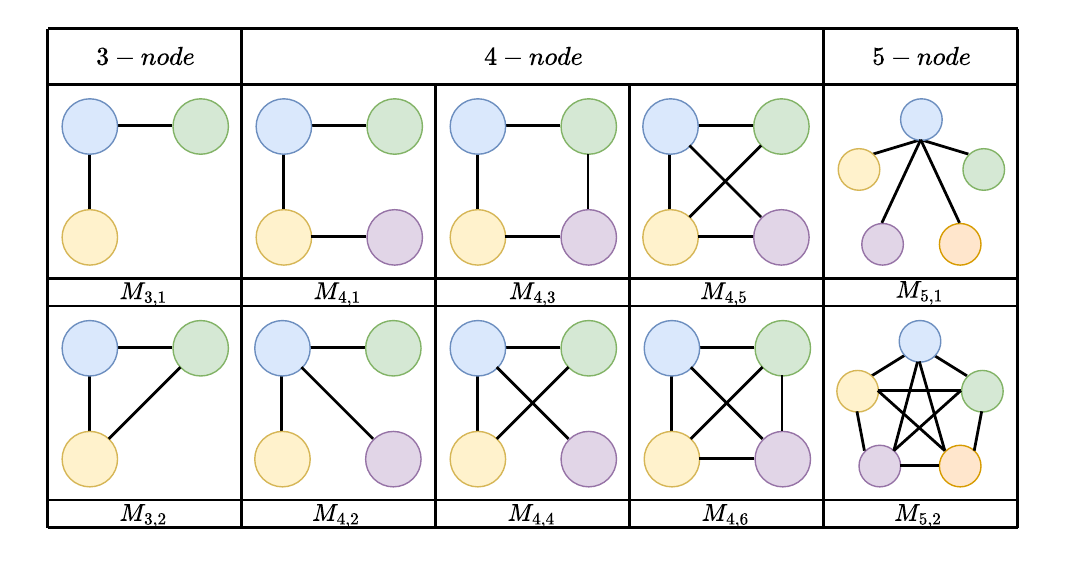} 
\caption{Network Motifs Employed in the Study.(Blue represents the target node, others represent neighbors.)}
\label{fig2}
\end{figure}

\subsubsection{Motifs Detection}
Each prototype is aligned with the local structure of a node within its category through prototypical projection, utilizing random walk sequences. These sequences are then reconstructed into a subgraph for each prototype, providing a subgraph representation. Building on this, we initiate motif discovery, where motifs are defined as statistically overrepresented substructures in networks \cite{milo2002network}. In a real network and $N$ random networks, a subgraph qualifies as a motif if it meets the following conditions, with a probability threshold $P_M$:
\begin{equation}
    p((f_{rand}(G_k) > f_{real}(G_k))) \leq P_M,
\end{equation}
where $f_{real}(G_k)$ is the frequency of a motif in real network $G_k$, and $f_{rand}(G_k)$ is the average frequency in all random networks about $G_k.$

The main advantage of motifs is the ability to capture the core connectivity of a network with fewer nodes and edges, maintaining a balance between complexity and interpretability. 
Too few nodes miss critical structural patterns, whereas too many reduce motif clarity. We focus on 3-node, 4-node, and selected 5-node motifs, as shown in Figure \ref{fig2}, to provide detailed insights into urban social segregation from a structural perspective. For each subgraph $G_k$ associated with a prototype $\mathbf{p}_k$, we count the frequency of the motif, forming the motif distribution $m_k$ for the current prototype.

\subsection{Urban Graph Structure Reconstruction}
The motif distribution derived from prototype vectors enables a thorough analysis of the spatial and mobility mechanisms influencing social segregation in cities. We use this distribution to guide the reconstruction of the urban graph. After establishing the motif distribution $\mathbf{M}$, we assign the corresponding motif distribution $m_i$ to each node, based on its local network structure $\mathcal{G}_i^{sub}$. 
With a reconstruction threshold  $\mathbf{\alpha}$, we incrementally adjust the adjacency matrix $A$ for nodes in a specific category. The steps are as follows:
\begin{equation}
    m_i^G \leftarrow \mathop{\arg\max}\limits_{m_k \in \mathbf{M}} ~~\mathrm{Sim}(\mathrm{Motif}(\mathcal{G}_i^{sub}), m_k),
\end{equation}
\begin{equation}
    A[i] = (1-\alpha\mathrm{KL})A[i] + (\alpha\mathrm{KL})A[tar],
\end{equation}
\begin{equation}
    \mathrm{KL} = KL(m_i^G, m_{tar}^G),
\end{equation}
\begin{equation}
A_{ij}^{\text{new}} = 
\begin{cases} 
1 & \text{if~}~A_{ij} > \beta \\
0 & \text{if~}~A_{ij} \leq \beta,
\end{cases}
\end{equation}
where $\mathcal{G}_i^{sub}$ denotes the subgraph of $\mathcal{G}$ with node $i$ and $A[i]$ denotes the $i$-th row of adjacency matrix. $KL(\cdot, \cdot)$ signifies the KL divergence, $\alpha$ represents the reconstruction weight factor, and $\beta$ stands for the edge creation threshold.

\subsection{Optimization}
During training, our objective is to understand the network structure underlying urban social segregation, using the social segregation index $\mathbf{d}^{SEG}$ as the ground truth. We enhance model's accuracy in predicting social segregation levels by minimizing the cross-entropy loss function $\mathcal{L}_{CrsEtp}$. In the prototype learning layer, we improve prototype interpretability by imposing constraints that refine their ability to capture key local structures. Following \citeauthor{zhang2022protgnn} \shortcite{zhang2022protgnn}, we incorporate a cluster loss $\mathcal{L}_{Clst}$, encouraging nodes to align more closely with their respective prototype vectors, and a separation loss $\mathcal{L}_{Sprt}$, which distances nodes from non-category prototypes. Additionally, in the motif detection module, we apply an encoding loss $\mathcal{L}_{Enc}$ to ensure the local subgraph encoder learns the optimal representation from subgraphs to prototype vectors. In summary, the loss functions we aim to optimize are as follows:
\begin{equation}
    \mathcal{L}_{CrsEtp} = \frac{1}{n}\sum_{i=1}^{N}\sum_{c=1}^{C}y_i^c\log(g_{\bf p}(h_i^c)),
\end{equation}
\begin{equation}
    \mathcal{L}_{Clst} = \frac{1}{n}\sum_{i=1}^{N} \mathop{\min}\limits_{j:\mathbf{p}_j \in \mathbf{P}_{y_j}} \parallel g_{\bf p}(h_i)-\mathbf{p}_j \parallel^2_2,
\end{equation}
\begin{equation}
    \mathcal{L}_{Sprt} = -\frac{1}{n}\sum_{i=1}^{N} \mathop{\min}\limits_{j:\mathbf{p}_j \notin \mathbf{P}_{y_j}} \parallel g_{\bf p}(h_i)-\mathbf{p}_j \parallel^2_2,
\end{equation}
\begin{equation}
    \mathcal{L}_{Enc} = \frac{1}{p}\sum_{k=1}^p \mathop{\min}\limits_{i:\mathcal{G}_{sub}^i \in \mathcal{G}_{sub}(y_i=k)} \parallel f_{sub}(\mathcal{G}_{sub}^i) - \mathbf{p}_k\parallel_2^2,
\end{equation}
where $N$ denotes the number of block nodes, $C$ represents the number of segregation levels, $y_i^c$ is the true label of node $i$, and $g_{\mathbf{p}}(\cdot)$ is the function of prototype layer. $\mathbf{P}_{y_j}$ represents the set of prototype vectors belonging to the same class as $y_i$, $p$ denotes the number of prototypes, $f_{sub}$ is the function of the subgraph encoder, and $\mathcal{G}_{sub}^i$ is the subgraph centered at node $i$. Combining these loss functions, the overall objective function is as follows:
\begin{equation}
    \mathcal{L} = \mathcal{L}_{CrsEtp} + \lambda_1\mathcal{L}_{Clst} + \lambda_2\mathcal{L}_{Sprt} + \lambda_3\mathcal{L}_{Enc}.
\end{equation}

\section{Experiments}

\subsubsection{Datasets} 
Beijing has evolved unique residential spatial distribution and mobility patterns during urban expansion, making it an ideal city for studying urban social segregation. 
We selected Beijing as the study city and divided it into 2104 blocks based on geographical divisions. To focus on social segregation in the urban core, we sampled 842 blocks within a 10-kilometer radius of the city center. The dataset includes a spatial graph $\mathcal{G}_s(\parallel\mathcal{V}\parallel=842,\parallel\mathcal{E}\parallel=6132)$ and an OD graph $\mathcal{G}_o(\parallel\mathcal{V}\parallel=842,\parallel\mathcal{E}\parallel=36334).$ Each block contains data on 21 categories of  POIs, hourly pedestrian flow indices, and 20 street view images for both summer and winter, obtained through uniform sampling within the block boundaries. 

\subsubsection{Baselines}
%\noindent\textbf{Baselines}~
In the experiments, we select GNN variants such as GCN \cite{kipf2016semi}, GAT \cite{velivckovic2017graph}, and GIN \cite{xu2018powerful}, along with interpretable models like ProtGNN\cite{zhang2022protgnn} and PGIB\cite{seo2024interpretable}. 
To better evaluate the improvements brought by graph prototype learning to GNN models, we also include GAT and GIN models augmented with the prototype framework. Through comparative analyses, we assess the effectiveness of our model in accurately predicting social segregation levels.

\subsubsection{Experiment Settings}

In the social segregation levels prediction task, we split the data into training, validation, and test sets with proportions of 0.6, 0.2, and 0.2, respectively. The social segregation index serves as the training label for the prototype network. A pre-trained ResNet50 \cite{he2016deep} is used to encode street view images, which are then combined with POIs and flow indices to generate 256-dimensional node attributes. These node attributes are subsequently mapped into a 128-dimensional latent space using a GNN module.
The learning parameters are set as follows: a learning rate of 0.001, a maximum of 3000 epochs, and a prototype projection interval of 50 epochs. The hyperparameters for the objective function are $\lambda_1=0.4, \lambda_2=0.2, \lambda_3=2.$ A crucial hyperparameter is the number of prototype vectors per class ($N_{proto}$), which is set to 5 based on preliminary experiments. These experiments show that five prototypes optimize both efficiency and interpretability. Using more prototypes increases training time, while fewer prototypes reduces motif discovery. Additionally, to ensure robustness, each experiment is repeated five times.

\subsection{Experimental Results}
\subsubsection{Social Segregation Levels Classification Task}
We assess the model's performance in predicting segregation levels. The results in Table \ref{table1} show that: 1) Compared to GCN, GAT, and GIN models, MotifGPL achieves the best performance in the prediction task. 
Furthermore, it offers a significant advantage in interpretability. While GNN models like GCN, GAT, and GIN are opaque box models without interpretability, which limits the investigation of social segregation, our graph prototype learning framework captures valuable information and improves interpretability through low-dimensional prototype vectors.
2) Compared to GNN interpretability methods like ProtGNN and PGIB, which are optimized for graph classification and underperform on node classification, our model achieves higher accuracy while maintaining interpretability. The prototype vectors in our model capture key factors related to social segregation, improve performance on node-level tasks, and support effective motif pattern discovery.

\begin{table}[t]
\centering
\begin{tabular}{ccc}
\toprule
      & \multicolumn{2}{c}{Segragation Index} \\
    \cmidrule(lr){2-3}
     & $Accuracy~(\uparrow)$ & $F1\!-\!score~(\uparrow)$ \\
    \midrule
    GCN & 0.7872$\pm$0.0002 & 0.7854$\pm$0.0002 \\
    GAT & 0.7692$\pm$0.0006 & 0.7686$\pm$0.0006 \\
    GIN & 0.7512$\pm$0.0008 & 0.7508$\pm$0.0008 \\
    ProtGNN & 0.7348$\pm$0.0014 & 0.7384$\pm$0.0016 \\
    PGIB & 0.7502$\pm$0.0010 & 0.7626$\pm$0.0008 \\
    GAT+Prototype & 0.7896$\pm$0.0001 & 0.7884$\pm$0.0001 \\
    GIN+Prototype & 0.7644$\pm$0.0006 & 0.7640$\pm$0.0005 \\
    {\bf MotifGPL} & {\bf 0.7990$\pm$0.0005}  & {\bf 0.7976$\pm$0.0006} \\
    \bottomrule
    \end{tabular}
%\caption{Evaluation of social segregation levels classification.}
\caption{Evaluation of social segregation classification.}
\label{table1}
\end{table}

\subsubsection{Motif Distribution Discovery}
\begin{figure}[t]
\centering
\includegraphics[width=0.9\columnwidth]{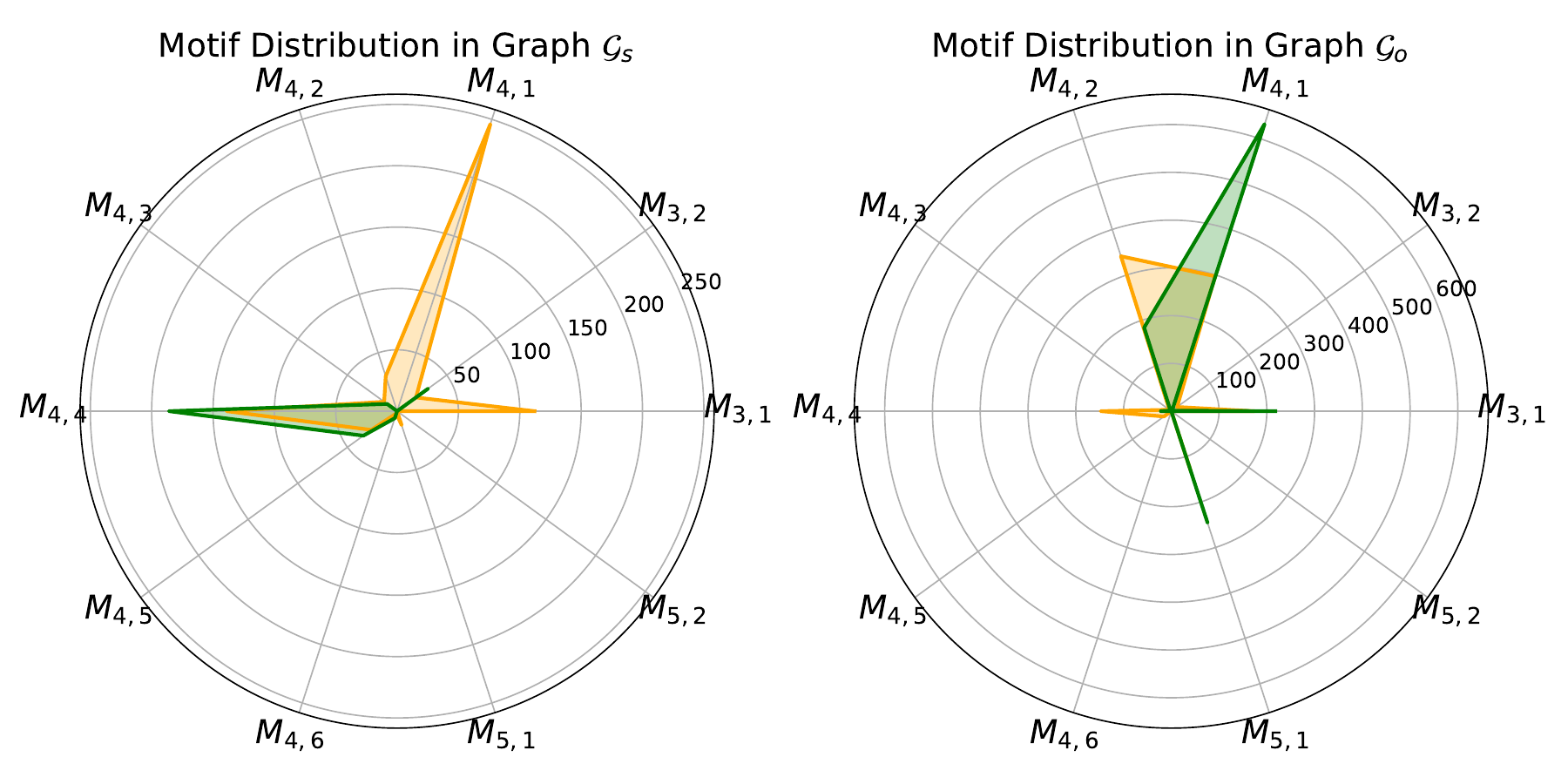} 
\caption{Motif Distribution of Prototypes in $\mathcal{G}_s$ and $\mathcal{G}_o.$}
\label{fig3}
\end{figure}

Next, we employ graph isomorphism algorithms to identify motif patterns of trained prototype. Figure \ref{fig3} shows the motif distribution corresponding to the prototypes of Graph $\mathcal{G}_s$ and $\mathcal{G}_o$, where green represents motifs of blocks with high segregation and orange indicates motifs of blocks with low segregation. 
In the spatial graph, motifs in high-segregation blocks are primarily found in $M_{4,4}$ and $M_{3,2}$. These circular motifs indicate that these blocks exhibit spatial clustering, forming enclosed community structures. In contrast, motifs in low-segregation blocks are concentrated in $M_{4,1}, M_{4,4},$ and $M_{3,1}$, with chain-like motifs suggesting a linear spatial distribution that enhances interaction with the urban environment.
In the OD graph, the motif distribution of blocks with varying segregation levels is primarily concentrated in chain-like ($M_{4,1}$ and $M_{3,1}$) and star-like ($M_{4,2}$ and $M_{5,1}$) patterns. Comparing the number of motifs in $M_{4,1}$, high-segregation blocks contain significantly more chain-like motifs than low-segregation blocks, indicating that residents in high-segregation blocks endure longer commutes. Additionally, star-like motifs in high-segregation blocks are more complex than those in low-segregation blocks, suggesting that residents in segregated blocks often travel to centralized blocks for daily activities, reflecting a clear separation between living and working spaces. Based on these findings, urban planners should consider shifting from traditional clustered housing to a linear distribution strategy when selecting sites for new affordable housing developments to mitigate segregation among residents. Additionally, improving amenities around high-segregation blocks will help residents more easily meet their daily living and working needs.

\begin{table}[t]
\centering
\begin{tabular}{ccccccc}
\toprule
         & $\alpha$ & $\beta$ & AEP & REP & UEP & Moran's I($\downarrow$)\\
\midrule

\multirow{4}{*}{$\mathcal{G}_s$}
    & \multicolumn{2}{c}{original} & 0.00 & 0.00 & 100.00 & 0.4159\\
    &  0.8 & 0.3 & 1.40 & 0.24 & 99.76 & 0.4043\\
    & 0.8 & 0.2 & 5.81 & 0.00 & 100.00 & 0.3751\\
    & 0.8 & 0.1 & 15.05 & 0.00 & 100.00 & 0.3169\\
\midrule
\multirow{4}{*}{$\mathcal{G}_o$} 
    & \multicolumn{2}{c}{original} & 0.00 & 0.00 & 100.00 & 0.2410\\
    & 0.8 & 0.3 & 1.89 & 0.32 & 0.9968 & 0.2341\\
    & 0.8 & 0.2 & 8.14 & 0.00 & 100.00 & 0.2118\\
    & 0.8 & 0.1 & 19.92& 0.00 & 100.00 & 0.1805 \\
    \bottomrule
    \end{tabular}
\caption{Results of urban graph reconstruction. AEP denotes added edge percentage, REP denotes removed edge percentage, and UEP denotes unchanged edge percentage.}
\label{table2}
\end{table}

\begin{figure}[t]
\centering
\includegraphics[width=0.9\columnwidth]{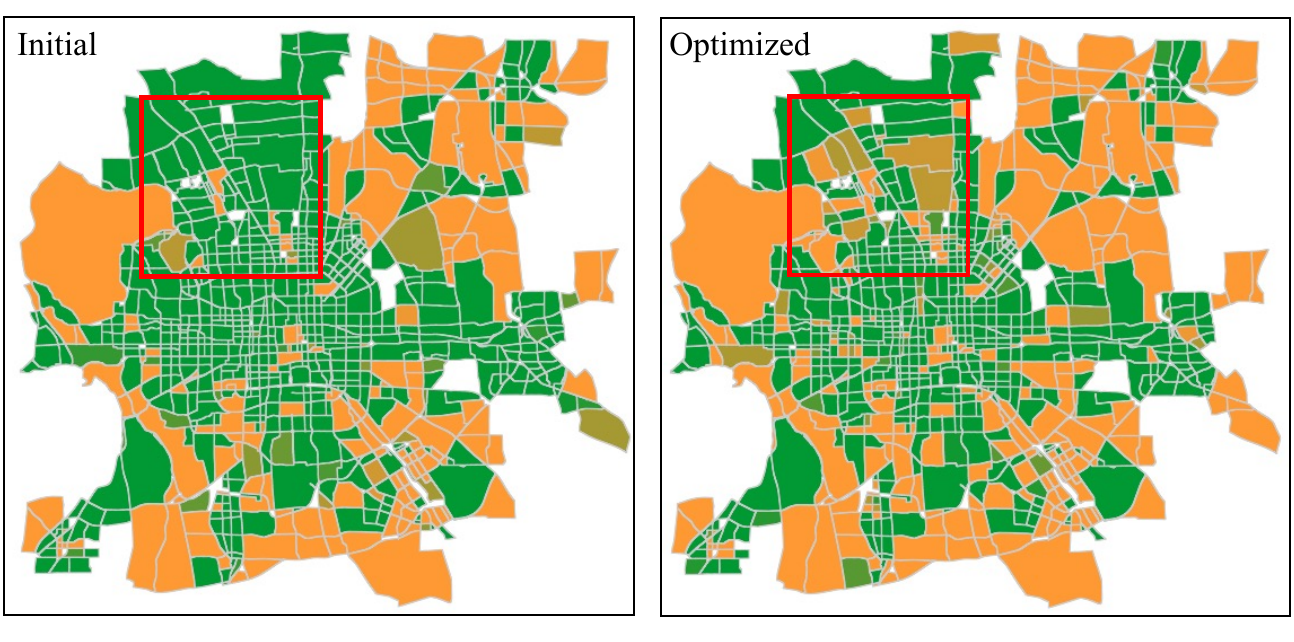} 
\caption{Comparison of Reconstruction Outcomes in Blocks Segregation Levels (Orange represents low segregation, green represents high segregation).}
\label{fig4}
\end{figure}

\subsubsection{Urban Graph Reconstruction}
Using the discovered motif distribution to guide the reconstruction of urban graph structures can facilitate minor modifications to the existing urban framework,ultimately reducing overall social segregation in the city.
According to \citeauthor{garreton2020exploring} \shortcite{garreton2020exploring}, we utilize Global Moran’s I to measure the overall degree of social segregation across Beijing, as it captures spatial autocorrelation and reflects dynamic changes in segregation patterns, unlike the static segregation index, which is unsuitable for dynamic network reconstruction.
Global Moran’s I ranges from -1 to 1, where a larger absolute value indicates higher spatial segregation. Table \ref{table2} presents the results of the urban graph structure reconstruction experiments. Here, $\alpha$ and $\beta$ represent the reconstruction weight and edge generation threshold, respectively. 
The results indicate that using motif distribution to guide spatial or OD graph reconstruction reduces social segregation.
At a low reconstruction level ($\alpha=0.8, \beta=0.3$), only minor changes (\textless 2.5\%) to the existing urban structure are required to decrease social segregation. 
Figure \ref{fig4} shows the changes in segregation levels in Beijing after urban graph reconstruction. Highlighted areas indicate that reconstruction can partially reduce social segregation.
In practice, enhancing connectivity between blocks by improving infrastructure or adding new transportation routes is a feasible approach for urban graph structure reconstruction. Experiments show that motif patterns provide novel insights for guiding reconstruction.

\subsubsection{Ablation Study}
\begin{table}[t]
\centering
\begin{tabular}{ccc}
\toprule
      & \multicolumn{2}{c}{Segragation Index} \\
    \cmidrule(lr){2-3}
     & $Accuracy~(\uparrow)$ & $F1\!-\!score~(\uparrow)$ \\
    \midrule
    {\bf MotifGPL} & {\bf 0.7990$\pm$0.0005}  & {\bf 0.7976$\pm$0.0006}\\
    w/o $\mathcal{G}_o$ & 0.7776$\pm$0.0005 & 0.7758$\pm$0.0005 \\
    w/o $\mathcal{G}_s$ & 0.7212$\pm$0.0002 & 0.7210$\pm$0.0002 \\
    w/o $X^{SV}$ & 0.7834$\pm$0.0008 
    & 0.7814$\pm$0.0009 \\
    w/o $X^{FL}$ & 0.7752$\pm$0.0005 
    & 0.7734$\pm$0.0006 \\
    w/o $X^{POI}$ & 0.7560$\pm$0.0003 
    & 0.7534$\pm$0.0004 \\
    \bottomrule
    \end{tabular}
\caption{Results of ablation study for MotifGPL.}
\label{table3}
\end{table}
We conduct ablation experiments on the classification task, focusing on graph structures and node attributes. As shown in Table \ref{table3}, the results indicate that the absence of either the spatial graph or the OD graph leads to a decrease in performance, suggesting that the local structures of these two graphs are crucial for analyzing urban social segregation from the perspective of motif patterns. Similarly, the absence of any node attribute leads to reduced performance, indicating that the selected urban attribute data reflect various aspects of segregation.

\section{Conclusion}
This paper presents a framework called Motif-Enhanced Graph Prototype Learning (MotifGPL), which integrates motif discovery with graph prototype learning to uncover insights related to social segregation within urban spatial structures and population movement patterns. We explore motif patterns associated with social segregation, providing a new perspective for addressing segregation issues in urban environments.
Our experimental results demonstrate that the motif patterns identified by the model have strong interpretability in real-world scenarios. This not only provides a novel methodological approach for investigating urban segregation but also offers substantial support for urban planning and development practices.

\section{Acknowledgments}

This work was supported by the National Natural Science Foundation of China (NSFC) (Grant No. 62106274) and the Fundamental Research Funds for the Central Universities, Renmin University of China (Grant No. 22XNKJ24).

% \bibliography{main}

\appendix
\newpage 
\section{Appendix}

\subsection{Datasets}
\subsubsection{Study Area}
\begin{figure}[ht]
  \centering
  \begin{subfigure}[b]{0.45\columnwidth}
    \includegraphics[width=\columnwidth]{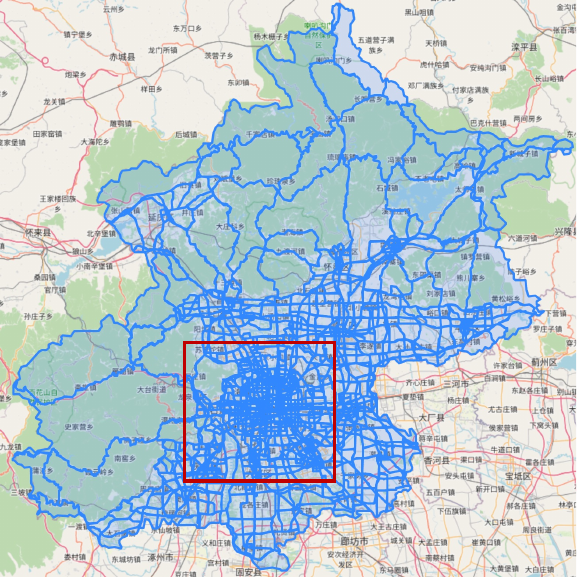}
    \caption{Beijing's overall administrative division.}
    \label{fig5:sub1}
  \end{subfigure}
  \hspace{5pt}
  \begin{subfigure}[b]{0.45\columnwidth}
    \includegraphics[width=\columnwidth]{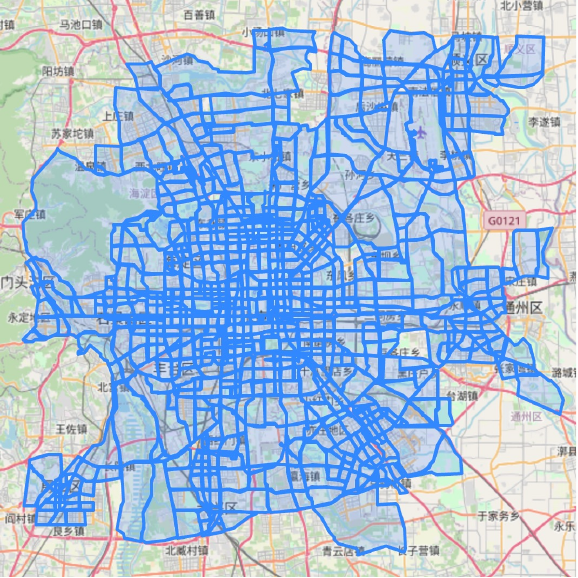}
    \caption{Central districts of beijing's blocks.}
    \label{fig5:sub2}
  \end{subfigure}
  \caption{Division of the study area.}
  \label{fig5}
\end{figure}

We select Beijing as the research area and subdivide the entire city into 2104 blocks based on the natural urban block planning, as detailed in Figure \ref{fig5:sub1}. Given the limited interaction between suburban residents and those in central Beijing, along with the difficulty in collecting data from suburban blocks, we focus primarily on the central urban districts of Beijing. 
The decision to focus on the central urban districts is driven by several key considerations: Firstly, this area has a high density of residential and commercial activities, providing a rich volume of data conducive to more comprehensive analysis. Secondly, as the core of political, economic, and cultural activities, the central urban districts play an important role in demonstrating and guiding the city's development patterns and urban planning. In the preliminary selection process, blocks with scarce data are systematically excluded to ensure that the study focused on areas where comprehensive and reliable data could be gathered. Ultimately, 842 blocks in the central urban districts are identified as the main study area, with their specific geographical locations shown in Figure \ref{fig5:sub2}. 

\subsubsection{Street View}
\begin{figure}[ht]
\centering
\includegraphics[width=0.9\columnwidth]{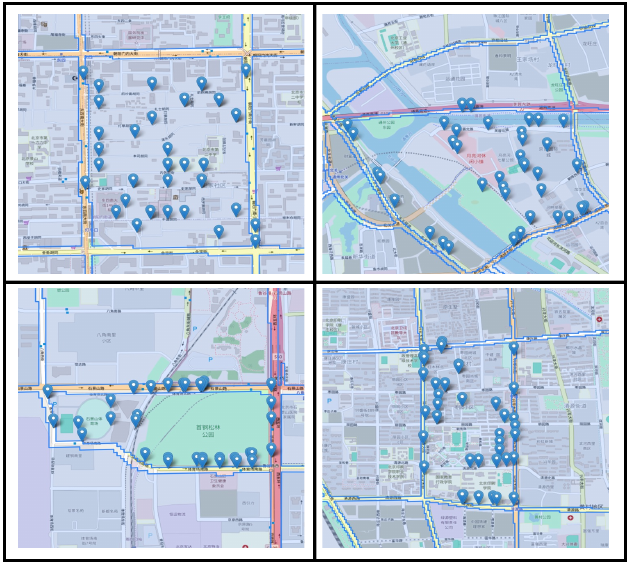} 
\caption{Street view selection points within each block.}
\label{fig6}
\end{figure}

\begin{figure}[ht]
\centering
\includegraphics[width=0.9\columnwidth]{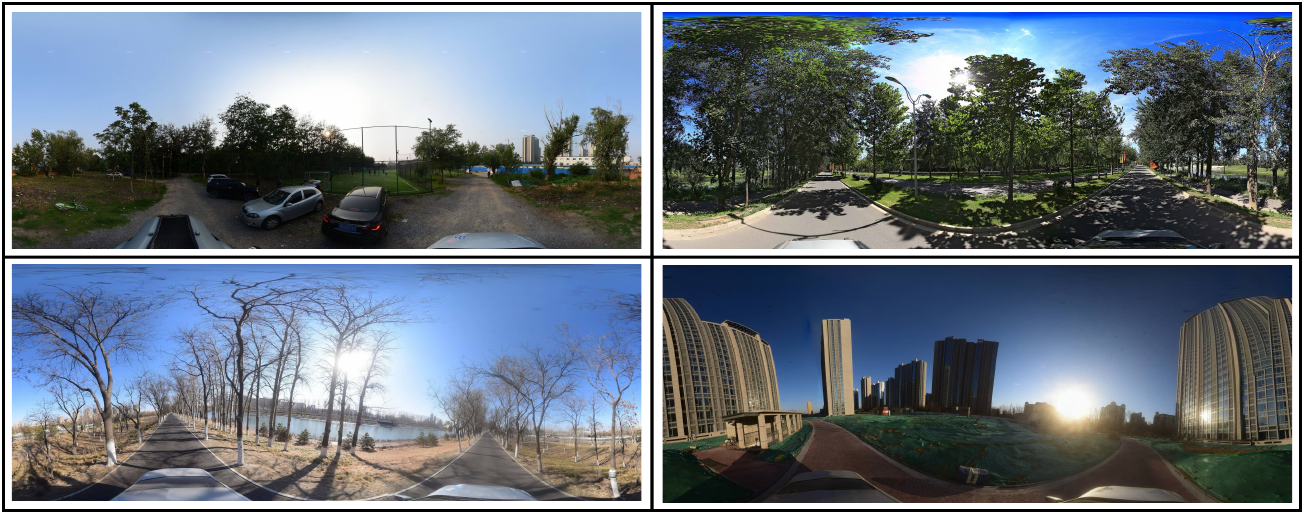} 
\caption{Examples of street view images in certain blocks.}
\label{fig7}
\end{figure}

To comprehensively exhibit the environmental characteristics of each urban block, we randomly collect 20 summer street view images and 20 winter street view images per block. To ensure that the collected street view data are broadly representative and thoroughly covered, distance constraints are implemented during the sampling process to ensure a uniform distribution of street view points and to avoid overlaps. Figure \ref{fig6} shows the specific locations of street view sampling points selected across some certain urban blocks in detail. Due to access restrictions in certain areas, it is not possible to directly obtain specific street view images from some internal parts of the blocks; hence, we choose few images from adjacent streets as substitutes. Street view images located on streets are assigned to the nearest urban block based on proximity. Figure \ref{fig7} displays the summer and winter street view images from different positions within the same block, illustrating not only the impact of seasonal changes on the visual environment of the blocks but also revealing the environmental diversity of urban landscapes within different positions of the area. 

\subsection{Experiments}
\subsubsection{Motif Distribution Discovery}
\begin{figure}[ht]
  \centering
  \begin{subfigure}[b]{0.4\columnwidth}
    \includegraphics[width=\columnwidth]{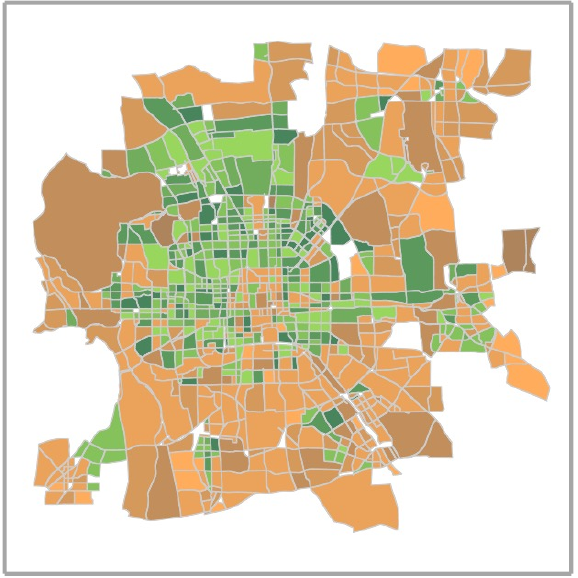}
    \caption{Motif global distribution map.}
    \label{fig8:sub1}
  \end{subfigure}
  \hspace{5pt}
  \begin{subfigure}[b]{0.4\columnwidth}
    \includegraphics[width=\columnwidth]{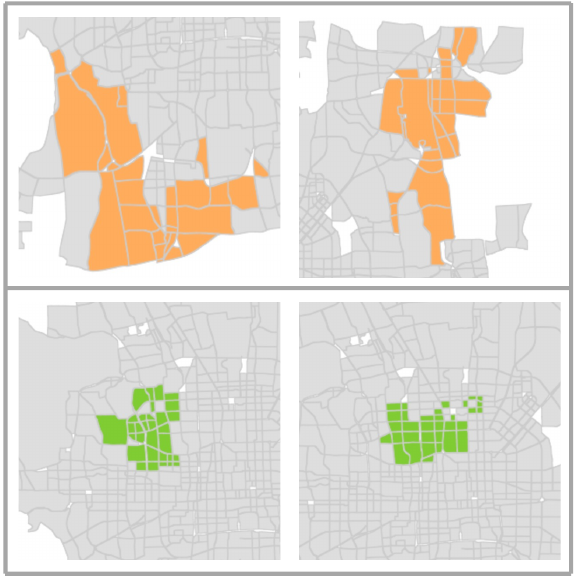}
    \caption{Local spatial structure map of specific motifs.}
    \label{fig8:sub2}
  \end{subfigure}
  \caption{Motif global distribution map and local spatial structure map of specific motifs.}
  \label{fig8}
\end{figure}

Figure \ref{fig8:sub1} shows the most similar motif patterns in each block within the motif distribution discovery module, where orange represents blocks with lower income segregation and green indicates higher income segregation. Figure \ref{fig8:sub2} shows local subgraph structures in $\mathcal{G}_s$ that closely match specific motifs. Observations reveal that lower income segregation blocks typically show a band-like distribution, whereas higher income segregation blocks exhibit centralized clustering. These patterns are confirmed by the motif analysis in Figure 4 of the main text.

\end{document}